# Forensic Video Steganalysis in Spatial Domain by Noise Residual Convolutional Neural Network

Mart Keizer[a,b], Zeno Geradts[a,c,*], Meike Kombrink[a,c]

[a]Netherlands Forensic Institute, The Hague, Netherlands
[b]Eindhoven University of Technology, Eindhoven, Netherlands
[c]University of Amsterdam, Amsterdam, Netherlands

## Abstract

This research evaluates a convolutional neural network (CNN) based approach to forensic video steganalysis. A video steganography dataset is created to train a CNN to conduct forensic steganalysis in the spatial domain. We use a noise residual convolutional neural network to detect embedded secrets since a steganographic embedding process will always result in the modification of pixel values in video frames. Experimental results show that the CNN-based approach can be an effective method for forensic video steganalysis and can reach a detection rate of 99.96%.

## 1. Introduction

Steganography is the hiding of secret information inside innocent looking (digital) objects, whereas, steganalysis is the science of detecting steganography (Li et al., 2011). This research relates to forensic steganalysis, which refers to a level of steganalysis that attempts to obtain further in-depth knowledge about a hidden message (Chutani and goyal, 2019). Steganography and (forensic) steganalysis have become an important field of interest in forensic science,

*Corresponding author.
*Email addresses:* `m.keizer@nfi.nl` (Mart Keizer), `z.geradts@nfi.nl` (Zeno Geradts), `m.kombrink@nfi.nl` (Meike Kombrink)

due to the rising popularity of steganography among criminals and terrorists (Choudhary, 2012)(European Commission, 2021). A reason for the rise in popularity is that "stand-alone" encryption methods have proved to be insecure. This is mainly a result of the successful and continuous effort of law enforcement agencies to compromise encrypted criminal communications, such as with the "EncroChat Hack" in 2020 (O'Rourke, 2020). Steganography can also be used along with encryption, which provides more security and robustness (Taha et al., 2019), because one first needs to find the encrypted secret and extract it from the carrier before it can be decrypted.

An object in which a secret is hidden is called a cover. Many covers can be used for steganography (e.g., text files, images, audio files), but in this paper, we focus on video steganography, i.e., steganography with video files as cover. This is a popular research area within steganography due to the high secret hiding potential of video files compared to, for example, images and audio files (Liu et al., 2019). Videos are in fact a moving stream of images and sounds and, therefore, any distortions in a video file might go by unobserved by humans. Additionally, video files are generally larger in size and, therefore, have more potential bytes to alter to embed a secret.

Criminals and terrorists can use video steganography to hide information and to communicate with each other (Garcia, 2018). There exist numerous real-world examples where steganography has been used with bad intentions (Dalal and Juneja, 2021). US officials claimed, for example, that Al-Qaeda used steganography to plan the 9/11 attack. Also in several cases, steganography was used to hide child pornography. Therefore, specialized tools need to be developed by law enforcement agencies to detect such criminal content on, for example, confiscated computers. Criminals can also communicate with video steganography by, for example, uploading an innocent looking video with an embedded message to a popular online video sharing platform. The intended receiver of the message only needs to know where to look to extract the video and its hidden content. This is very difficult to detect because of the vast number of video files on such platforms. Unfortunately, the research on video steganalysis is lacking behind the research on video steganography. Therefore, forensic video steganalysis requires more interest from forensic researchers.

One possible approach for forensic steganalysis is to use a convolutional neural network (CNN). A CNN is a special type of artificial neural network that is most commonly applied to analyze images, therefore one can use a CNN to conduct forensic steganalysis from the perspective of the spatial domain, where the modification of pixel values are analyzed to detect a hidden secret.

This study evaluates a CNN and its video steganalysis capabilities (i.e., its ability to detect video steganography) in spatial domain. The Noise Residual Convolutional Neural Network (NR-CNN), proposed by (Liu and Li, 2020), is trained on a video steganography dataset to detect two steganography tools: Deep Video Steganography (Sathyan, 2019) and MSU StegoVideo (Moscow State University, 2006). The remainder of this paper is organized as follows. Section 2 describes the related research papers from this research. Section 3 describes how the steganography dataset is created and how the NR-CNN is trained and evaluated. The results are presented and discussed in section 4 and conclusions and future directions are in section 5.

2. Related work

*2.1. Deep Steganography*

Deep Steganography refers to steganography implemented with a deep neural network, which is an artificial neural network with multiple layers between the input and output layers. A Deep Steganography method was proposed by (Baluja, 2017) to hide an image inside an other image of the same size. The proposed Deep Steganography network consists of three parts: a preparation network, a hiding network and a reveal network, which are trained together on an image dataset. The input of the network is a cover image and a secret image and the output is the target cover image and the target secret image after the embedding and extracting process. During training the difference between the input and the output images are used as error terms. In this way, the network can learn how to hide images inside each other with the lowest number of noticeable differences. Once the network is trained, it is split into an encoding and decoding network, which can then be used as a image steganography tool. In addition, since videos basically consists of a number of combined images, this tool can also be transformed into a video steganography tool.

## 2.2. MSU StegoVideo

MSU StegoVideo[1] is a video steganographic tool developed by (Moscow State University, 2006). The tool can be used to hide any file inside a video, with small video distortions. It has two important settings which are the data redundancy, which determines the amount of data that is hidden in each frame, and a noise setting, which determines the amount of added noise. The tool also requires a password to embed and extract the hidden secret. The Moscow State University does not provide any source code for MSU StegoVideo.

## 2.3. Noise Residual Convolutional Neural Network

A universal steganalysis method was proposed by (Liu and Li, 2020) to detect both intraprediction mode and motion vector based steganography. Since, both steganography methods eventually lead to the modification of pixel values in decoded frames, they designed a Noise Residual Convolutional Neural Network (NR-CNN) from the perspective of the spatial domain. In a data-driven manner the network automatically learned features and implemented classification. The trained NR-CNN reached a detection accuracy from 59.82% up to 99.74% for different embedding rates.

---

[1] https://www.compression.ru/video/stego_video/index_en.html

## 3. Method

A video dataset is created with a mix of regular videos and two types of steganography videos. The videos from that dataset along with their corresponding class labels are used to train a convolutional neural network. Specifically, it is trained to classify frames from the videos into three categories: regular, deep-stego and MSU-stego. Thus, the trained network should be able to determine if a video is a regular video or a video that is embedded with one of the two steganographic embedding techniques.

### 3.1. Data

The VLOG dataset (Fouhey et al., 2018) is used to obtain video material for the experiments. This dataset consists of 2.747 video files with a total duration of 314 hours and over 30 million frames, therefore, we do not require the whole dataset for our experiment. However, we need to generalize our dataset as much as possible, to make sure the trained network works well on unseen data. Therefore, the final dataset should contain video frames from as much different videos as possible. Hence, the VLOG dataset is transformed into a "10-sec" dataset by splitting the videos into videos of around 10 seconds each. Now, when we randomly pick a video from the 10-sec dataset we obtain a random section from a random video from the VLOG dataset. Next, we define a parameter *k*, which represents the number of 10-second videos we want in the final steganography dataset (stego-dataset). This stego-dataset, therefore, will contain *k*/3 Deep Steganography (DVS) videos, *k*/3 MSU StegoVideo (MSU) videos, and *k*/3 regular videos. The regular videos are created by simply copying *k*/3 videos from the 10-sec dataset into the stego-dataset and removing them from the 10-sec dataset.

#### 3.1.1. Deep Video Steganography dataset

The deep neural network hiding technique, proposed by (Baluja, 2017) and discussed in section 2.1, embeds an image into another image of the same size. In figure 1 an example is shown. This technique was implemented by (Sathyan, [2019](#)) for usage with video frames and made publicly available in the Deep Video Steganography (DVS) repository[2]. This repository is used to create the DVS videos for the stego-dataset, which is done by picking two distinct random videos from the 10-sec dataset and hiding one of those videos inside the other using DVS. The resulting video is added to the stego-dataset and the two randomly picked videos are removed from the 10-sec dataset. We repeat this $k/3$ times.

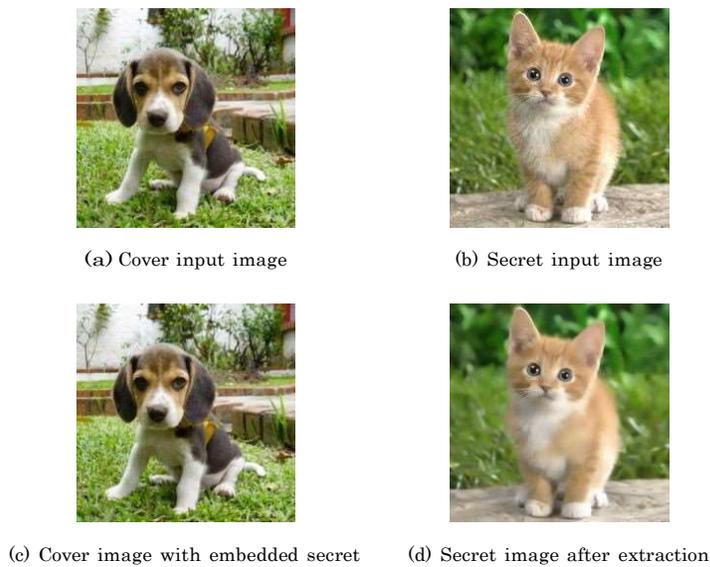

(a) Cover input image  (b) Secret input image

(c) Cover image with embedded secret  (d) Secret image after extraction

Figure 1: Example of the deep neural network hiding technique, with a picture of a cat as secret image and a picture of a dog as cover image.

Although the DVS technique is visually difficult to detect, the number of bits required to encode a secret frame inside a cover frame of the same size is between 1 and 4 bits per pixel (bpp). Given that previous studies (Xuan et al., 2005)(Zou et al., 2006) have demonstrated that bit rates of 0.1 bbp can already be discovered by statistical analysis, it is expected that the DVS videos can also be detected by a CNN-based approach.

---

[2]https://github.com/anilsathyan7/Deep-Video-Steganography-Hiding-Videos-in-Plain-Sight

### 3.1.2. MSU StegoVideo dataset

MSU StegoVideo[3] (Moscow State University, 2006), discussed in section 2.2, is a public video steganographic tool that can hide a secret file inside the frames of a video. Figure 2 shows an example of a secret text message embedded into a frame from a landscape video. We use MSU StegoVideo to create the MSU videos for the stego-dataset. This is done by randomly picking and concatenating *k*/3 distinct videos from the 10-sec dataset into a large video file. Next, we hide a text file containing Lorem Ipsum text inside the concatenated video using MSU StegoVideo, with data redundancy set to 2 and noise set to 100. The video with the embedded secret is then split into 10-second videos again, now all containing a part of the secret.

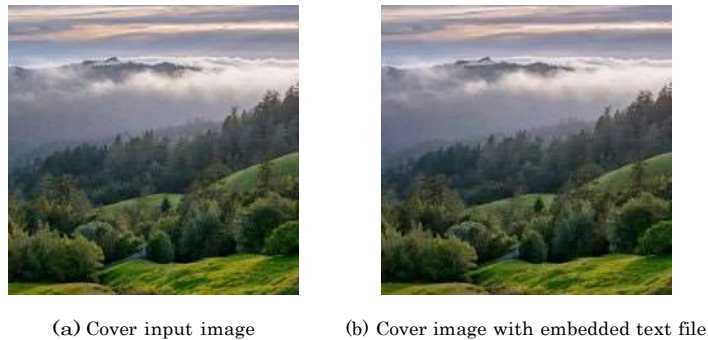

(a) Cover input image     (b) Cover image with embedded text file

Figure 2: Example of the MSU StegoVideo tool, with a frame from a landscape video as cover.

MSU StegoVideo is not open source and is, therefore, difficult to analyze. However, we approximated that the embedding rate of the secret text file is 0.1 bits per pixel (bpp), which is at least 10 times lower compared to the DVS technique. This might result in a lower detection rate compared to DVS.

---

[3]https://www.compression.ru/video/stego_video/index_en.html

| #  | Layer                       | Input            | Output        | Kernel Size | Notes                                  |
|----|-----------------------------|------------------|---------------|-------------|----------------------------------------|
| 1  | Residual convolutional layer| 1 channel        | 34 channels   | (5*5)       | Stride = 1, Batch Normalization, PTLU  |
| 2  | Convolutional layer 1       | 34 channels      | 34 channels   | (3*3)       | Stride = 1, Batch Normalization, ReLu  |
| 3  | Convolutional layer 2       | 34 channels      | 34 channels   | (3*3)       | Stride = 1, Batch Normalization, ReLu  |
| 4  | Convolutional layer 3       | 34 channels      | 34 channels   | (3*3)       | Stride = 1, Batch Normalization, ReLu  |
| 5  | Average pooling layer 1     | 34 channels      | 34 channels   | (2*2)       | Stride = 2                             |
| 6  | Steganalysis residual block 1| 34 channels     | 34 channels   | (3*3)       |                                        |
| 7  | Average pooling layer 2     | 34 channels      | 34 channels   | (3*3)       | Stride = 2                             |
| 8  | Steganalysis residual block 2| 34 channels     | 34 channels   | (3*3)       |                                        |
| 9  | Average pooling layer 3     | 34 channels      | 34 channels   | (3*3)       | Stride = 2                             |
| 10 | Convolutional layer 4       | 34 channels      | 32 channels   | (3*3)       | Stride = 1, Batch Normalization, ReLu  |
| 11 | Average pooling layer 4     | 32 channels      | 32 channels   | (2*2)       | Stride = 2                             |
| 12 | Convolutional layer 5       | 32 channels      | 16 channels   | (3*3)       | Stride = 1, Batch Normalization, ReLu  |
| 13 | Convolutional layer 6       | 16 channels      | 16 channels   | (3*3)       | Stride = 3, Batch Normalization, ReLu  |
| 14 | Fully connected layer       | 16*3*3 features  | 3 features    | NA          |                                        |
| 15 | Softmax layer               | 3 features       | 3 outcomes    | NA          | 0: regular, 1: deep stego, 2: MSU stego|

Table 1: The architecture of the Noise Residual Convolutional Neural Network (Liu and Li, 2020). The first layer computes the residual, the twelve subsequent layers perform the feature extraction and the final two layers complete the classification.

*3.2. Network Architecture*

We chose to use the Noise Residual Convolutional Neural Network (NR-CNN), as proposed by (Liu and Li, 2020) and discussed in section 2.3. They showed that the NR-CNN can detect both intraprediction mode and motion vector-based steganography by detecting the modification of pixel values in decoded frames. The steganography methods we aim to detect are not based on intraprediction modes or motion vectors but do also modify pixel values, therefore, the prospect is that the NR-CNN will also be effective in detecting both the DVS and MSU embedding approach.

The network architecture of the NR-CNN is shown in figure ??. The first layer is the residual convolutional layer "ResConv" whose role is to compute the steganographic noise residual from the video frames. It consists of 34 custom-made filters showed in figure 3: 30 high-pass filters and 4 global filters. These filters will be automatically optimized during training. The input of the "ResConv" layer is single channel 224x224 image data (gray-scaled video frames). The feature extraction part of the architecture consists of 6 convolutional layers with batch normalization, 4 average pooling layers and 2 steganalysis residual blocks. The steganalysis residual blocks are based on ResNet (He et al., 2016), but are improved for steganalysis by changing the final mapping from $F(x) + x$ to $x - F(x)$ to retain the steganographic residual signal. The classification part of the network consists of a fully connected (linear) layer and a softmax layer with three outputs representing the three possible classes: regular, deep-stego and MSU-stego.

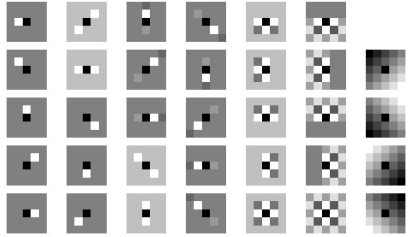

Figure 3: The 34 custom-made filters from the residual convolution layer of which the first 30 are high-pass filters and the last 4 are global filters. The filters will be optimized during training.

3.3. Experimental setup

The final steganography dataset contains $k = 3555$ videos, which are split into a training set, a validation set and a test set with respectively 1700, 174, and 1681 videos. The training set is used to train the NR-CNN and the validation set is used to validate the network during training. The test set is used to validate the network after training.

Before training, the videos from the training set are split into 5 segments, i.e., 2-second videos. From each segment, a random frame is picked to train the network with. This avoids training the network on possibly nearly similar subsequent frames, which may result in a network that is only working on the training set, i.e. overfitting. The frames are resized to 224x224 pixels and transformed into single channel gray-scale images. During the final evaluation of the network, we have a similar pre-processing step, but the videos from the test set are split into 20 segments instead of 5 to eventually evaluate the trained network on more frames.

The NR-CNN is implemented on the deep learning framework PyTorch, and is trained on 8,000 frames with a batch size of 20. The rest of the training setup is similar to the setup used by (Liu and Li, 2020) for the NR-CNN. Hence, the number of training iterations is 150 epochs, and the network optimizer is AdaDelta with learning rate $lr$ = 0.4, decay rate $\rho$ = 0.95, a weight decay of $5 * 10^{-4}$, and epsilon parameter $eps$ = $1 * 10^{-8}$. During training we try minimizing the Cross-Entropy loss.

The complete source code is available on Github[4].

4. Results & Discussion

In this section, we will evaluate the effectiveness of the trained NR-CNN on the detection of the Deep Video Steganography and MSU StegoVideo tools. The results are shown in table 2 and 3, which lists the predictions from the network on, respectively, the frames of the training set and the test set. The training set with 34,000 frames has an accuracy of 99.99% with 3 wrong predictions, and the test set with 33,620 frames has an accuracy of 99.96% with 13 wrong predictions. During inspection of the frames that where wrongly predicted we found a few possible causes for the faulty classifications. Out of the 16 frames there where 6 almost completely black frames, which may indicate that black frames are difficult to classify. Also 6 frames with content from a CCTV camera where wrongly classified, probably due to the visible horizontal lines that we typically find in CCTV camera footage. This conjecture is supported by two other wrongly predicted frames, which also had horizontal lines in the frames

---

[4]https://github.com/mjbkeizer/Video-Steganalysis-in-Spatial-Domain-by-Neural-Network

content. The two remaining frames did not show any clear possible cause of misclassification.

| Train labels \ Predictions | regular | deep-stego | MSU-stego |
|---|---|---|---|
| regular | 11580 | 0 | 0 |
| deep-stego | 1 | 10817 | 2 |
| MSU-stego | 0 | 0 | 11600 |

Table 2: The predictions from the NR-CNN per label class of the training set. The values represent the number of frames.

| Test labels \ Predictions | regular | deep-stego | MSU-stego |
|---|---|---|---|
| regular | 11620 | 0 | 0 |
| deep-stego | 2 | 10512 | 6 |
| MSU-stego | 0 | 5 | 11475 |

Table 3: The predictions from the NR-CNN per label class of the test set. The values represent the number of frames.

These results show that the NR-CNN is able to detect both the DVS and MSU embedding approaches with a remarkably high accuracy. One of the reasons for these high detection rates could be the fact that both steganography techniques leave behind some sort of clearly distinguishable watermark on the video frames. Such watermarks might be relatively easy to detect for convolutional neural networks, since CNN's are designed to classify shapes that have a number of similar features, which is the case for watermarks. It is however expected that every video steganography method will leave behind some sort of watermark and that makes the CNN-based steganalysis approach very promising. Another reason for the high accuracy on both techniques could be the high embedding rate, which is between 1 and 4 bits per pixel for DVS and approximately 0.1 for the MSU technique. However, if the embedding rate would have a significant impact on the detection rate, we should have expected to obtain more faulty predictions for the MSU-stego class than for the deep-stego class, which is not the case.

5. Conclusions & future directions

The high accuracy of 99.99% and 99.96% on the train and test set, respectively, show that the Noise Residual Convolutional Network (NR-CNN), from (Liu and Li, 2020), is not limited to detecting intraprediction mode and motion vector-based steganography, but can also be trained to accurately detect and classify the Deep Video Steganography (DVS) and MSU StegoVideo (MSU) methods. Given that the NR-CNN is also able to detect the DVS and MSU steganographic methods, which are completely different in their embedding techniques compared to the intraprediction mode and motion vector-based methods, we can start considering the NR-CNN as a promising general forensic video steganalysis network. However, it should be noted that, for now, the detection capabilities of the NR-CNN are only tested on steganography methods that where present in the training set. Future studies should, therefore, investigate whether the NR-CNN is also able to detect steganography methods that it has not seen before.

The resulting trained NR-CNN from this research can be used to accurately identify and classify steganography videos created with the Deep Video Steganography and the MSU StegoVideo tool. Furthermore, the created dataset could be expanded to train the network in detecting and classifying additional steganographic methods. However, in order to create the videos for the dataset, the corresponding tools for these steganographic methods have to be available. This is the case for publicly available steganography methods but it is, for example, also possible that a new steganographic tool is found on a confiscated computer from a person of interest. This tool could then be utilized to expand the created steganography dataset, by creating additional steganography videos with a video dataset and self fabricated hidden secrets. Subsequently, the network has to be retrained on the expanded dataset to also detect and classify that new tool in the future.

Experimental results show that a convolutional neural network based approach to detect and classify video steganography can perform exceptionally well. Future research should investigate whether the NR-CNN can be trained to use as a general forensic steganalysis tool.